\pdfoutput=1
\documentclass[10pt,twocolumn,letterpaper]{article}

\usepackage[pagenumbers]{cvpr} 

\usepackage[utf8]{inputenc} 
\usepackage[T1]{fontenc}    
\usepackage{microtype}      
\usepackage{xcolor,colortbl}         
\usepackage{xspace}
\usepackage{amsmath,amssymb,amsfonts,dsfont,pifont,bm,bbm,mathrsfs,mathtools,nicefrac}
\usepackage{algorithm,algorithmicx,algpseudocode,listings}
\usepackage{wrapfig}
\usepackage{booktabs,multirow,multicol,adjustbox,diagbox,threeparttable}
\usepackage{newfloat,graphicx}

\definecolor{citeblue}{HTML}{0071bc}
\definecolor{textpurple}{RGB}{135,89,201}
\definecolor{Gray}{gray}{0.90}
\newcolumntype{g}{>{\columncolor{Gray}}c}
\definecolor{ffe1da}{RGB}{255,225,218}
\definecolor{F7E0D5}{RGB}{247,224,213}
\definecolor{darkF7E0D5}{RGB}{209,154,128}

\usepackage[pagebackref=true,breaklinks=true,colorlinks=true,citecolor=citeblue,bookmarks=false]{hyperref}
\usepackage[capitalize]{cleveref}  
\usepackage{bbding}

\newcommand{\method}{SCFS\xspace}
\newcommand{\vm}[1]{\mbox{\boldmath $#1$}}

\newcommand{\authorskip}{\hspace{2.5mm}}

\definecolor{codegreen}{rgb}{0,0.6,0}
\definecolor{codegray}{rgb}{0.5,0.5,0.5}
\definecolor{codepurple}{rgb}{0.58,0,0.82}
\definecolor{backcolour}{rgb}{1.0,1.0,1.0}
\lstdefinestyle{mystyle}{
    backgroundcolor=\color{backcolour},
    commentstyle=\color{codegreen},
    keywordstyle=\color{magenta},
    numberstyle=\tiny\color{codegray},
    stringstyle=\color{codepurple},
    basicstyle=\ttfamily\scriptsize,
    breakatwhitespace=false,
    breaklines=true,
    captionpos=b,
    keepspaces=true,
    showspaces=false,
    showstringspaces=false,
    showtabs=false,
    tabsize=2
}
\crefname{section}{Sec.}{Secs.}
\Crefname{section}{Section}{Sections}
\Crefname{table}{Table}{Tables}
\crefname{table}{Tab.}{Tabs.}
\crefname{figure}{Fig.}{Figs.}
\Crefname{figure}{Figure}{Figures}
\crefname{equation}{Eq.}{Eqs.}

\def\cvprPaperID{..} 
\def\confName{..}
\def\confYear{..}

\title{Semantics-Consistent Feature Search for \\ Self-Supervised Visual Representation Learning}

\author{
    Kaiyou Song \authorskip \ 
    Shan Zhang \authorskip \ 
    Zihao An \authorskip \
    Zimeng Luo \authorskip \ 
    Tong Wang \authorskip \ 
    Jin Xie \authorskip \ 
    \\[3pt]
    MEGVII Technology \\[3pt]
    \small{\texttt{\{songkaiyou, zhangshan, anzihao, luozimeng, wangtong, xiejin\}@megvii.com}}
}

\begin{document}

\maketitle
\begin{abstract}

In contrastive self-supervised learning, the common way to learn discriminative representation is to pull different augmented ``views'' of the same image closer while pushing all other images further apart, which has been proven to be effective.
However, it is unavoidable to construct undesirable views containing different semantic concepts during the augmentation procedure.
It would damage the semantic consistency of representation to pull these augmentations closer in the feature space indiscriminately.
In this study, we introduce feature-level augmentation and propose a novel semantics-consistent feature search (\method) method to mitigate this negative effect.
The main idea of \method is to adaptively search semantics-consistent features to enhance the contrast between semantics-consistent regions in different augmentations.
Thus, the trained model can learn to focus on meaningful object regions, improving the semantic representation ability.
Extensive experiments conducted on different datasets and tasks demonstrate that \method effectively improves the performance of self-supervised learning and achieves state-of-the-art performance on different downstream tasks.

\end{abstract}
\section{Introduction}
\label{sec:intro}

Due to the tremendous potential in learning discriminative feature representation without using data annotations, self-supervised learning has received much attention in the representation learning field.
Contrastive learning \cite{moco_2020,simclr_2020}, as a type of discriminative self-supervised learning method, is heavily studied and has shown remarkable progress in the computer vision field in recent years.
It aims at pulling different augmented ``views'' of the same image (positive pairs) closer while pushing diverse images (negative pairs) far from each other.
To this end, a contrastive loss between the features of different views extracted from an encoder network is employed to train the encoder network end-to-end.
According to whether the negative pairs are used, current contrastive learning can be generally divided into two categories.

\begin{figure}[t]
\centering
\includegraphics[width=0.95\columnwidth]{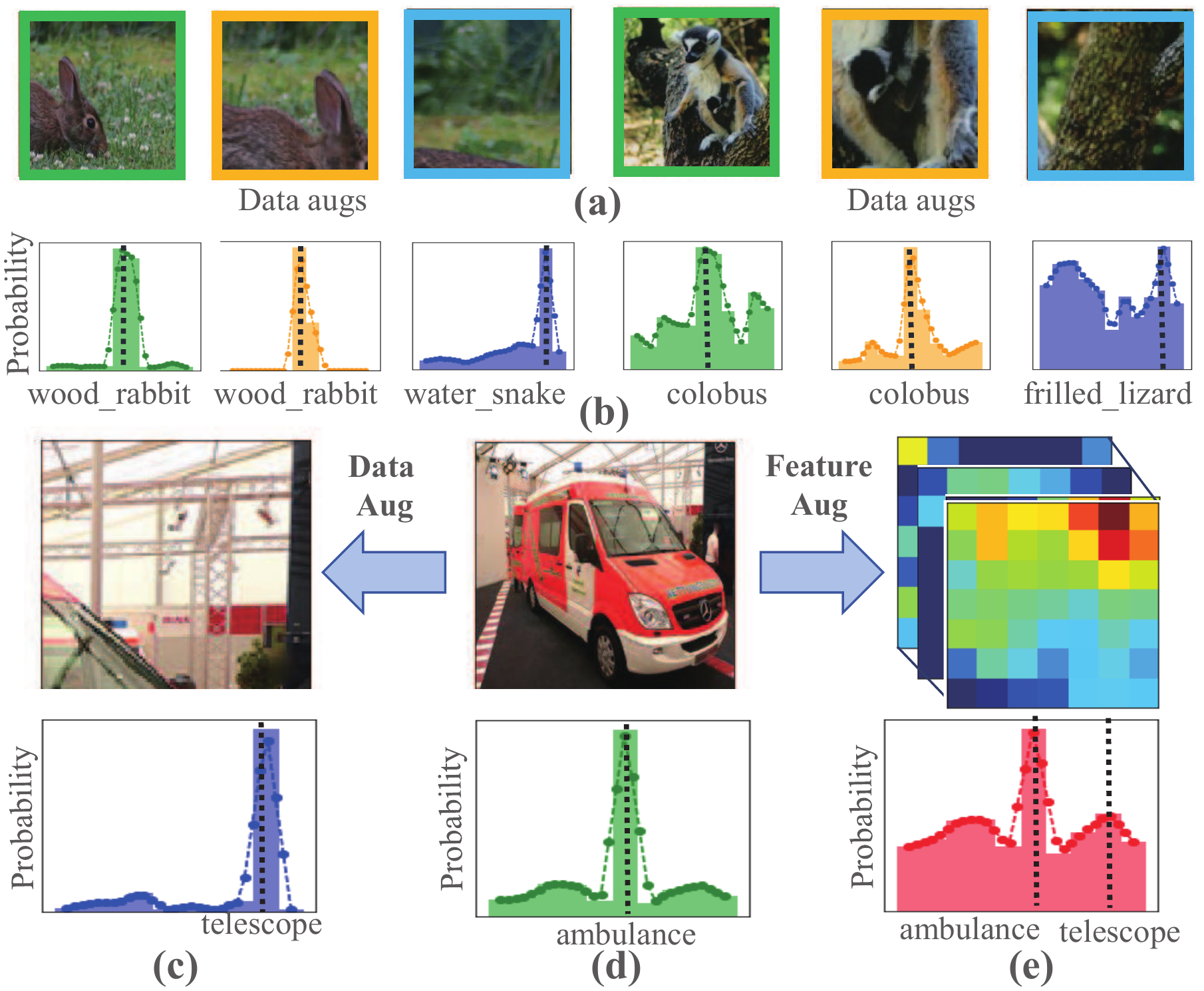}
\caption{Semantic inconsistency of over-augmentation.
(a) shows three augmentations of two images, in which the third augmentation is over-augmented and contains only background.
(b) shows category probability distributions of the corresponding images in (a), which are obtained from a supervised pre-trained ResNet50 \cite{resnet_2016} model.
(c)(d)(e) show three different samples of an image (the data-augmented image, the original image, and the semantics-consistent feature-augmented sample generated by Eq. \ref{equ:attention} in this study) and their corresponding probability distributions, which point out that the over-augmented image generates different category with the original image, while the feature-augmented sample gets a balanced category probability.}
\label{fig:introduction}
\end{figure}

The first category \cite{moco_2020,simclr_2020} utilizes both positive pairs and negative pairs for contrast.
MoCo \cite{moco_2020,mocov2_2020} uses a momentum update mechanism to maintain a memory bank of negative examples.
SimCLR \cite{simclr_2020,simclrv2_2020} directly trains a single encoder network with a large batch size to ensure sufficient positive and negative samples for learning. 
Based on MoCo and SimCLR, some methods \cite{momentum2tea_2021,msf_2021,isd_2021,nnclr_2021,dcl_2021,clsa_2021,adco_2021,mocov3_2021,hsa_2022,hcsc_2022} are proposed to improve the performance.
For example, MSF \cite{msf_2021}, ISD \cite{isd_2021} and NNCLR \cite{nnclr_2021} aim to search semantics-consistent samples for contrast, solving the false negative problem.
While some studies, such as Momentum2Teacher \cite{momentum2tea_2021} and DCL \cite{dcl_2021}, aim to solve the limitation that large batch size is necessary for satisfactory performance.

The second category of contrastive learning methods \cite{byol_2020,simsiam_2021,swav_2020,barlowtwins_2021,vicreg_2021,obow_2021,dino_2021,crafting_2022} only constructs positive pairs for contrast.
Based on MoCo \cite{moco_2020} and SimCLR \cite{simclr_2020}, respectively, BYOL \cite{byol_2020} and SimSiam \cite{simsiam_2021} abandon the negative samples and use an asymmetric architecture to avoid model collapse.
SwAV \cite{swav_2020} uses online clustering to cluster samples and forces the consistency among cluster assignments of different augmentations. 
After that, some studies \cite{dino_2021,clsa_2021,nnclr_2021,adco_2021} point out that enriching the augmented samples can improve the performance of contrastive learning.
In addition, the study in \cite{crafting_2022} shows that improving the quality of positive augmented samples is important for self-supervised learning.

However, it is unavoidable to construct data augmentations containing different semantic concepts.
\cref{fig:introduction}(a) shows three augmentations of two images, in which the third augmentation is over-augmented and contains only the background.
\cref{fig:introduction}(b) shows category probability distributions of the corresponding images in (a), which are obtained from a supervised pre-trained ResNet50 \cite{resnet_2016}.
We observed that the probability distribution of over-augmented images changes greatly compared with the first two augmentations, which indicates that the semantic information of the over-augmented images deviates from the normally-augmented images.
Similar observation can be found in \cref{fig:introduction}(c)(d).
The original image in (d) shows a max probability for ``ambulance'', while the over-augmented image in (c) represents the different category ``telescope''.
Due to such semantic inconsistency, conducting contrastive learning on these over-augmentations is harmful to representation learning.
In this study, we found that semantics-consistent feature augmentation (\cref{fig:introduction}(e), generated by \cref{equ:attention}) can balance the original semantics ``ambulance'' and the over-augmented semantics ``telescope'', which can alleviate the influence of semantic inconsistency.

Motivated by this observation, we propose a novel semantics-consistent feature search (\method) method to alleviate the negative influence of semantic inconsistency in contrastive learning.
\method utilizes the global feature of a view to adaptively search the semantics-consistent features of another view for contrast according to their similarity.
It constructs informative feature augmentations and conducts contrast learning between feature augmentations and data augmentations.
Thus, the pre-trained model can learn to focus on meaningful object regions to alleviate the negative influence of unmatched semantic alignment in current contrastive learning for better representation learning.
In addition, the feature search is conducted on multiple layers of the backbone network, further enhancing the semantic alignment at different scales of features.
Extensive experiments conducted on different datasets and tasks demonstrate that \method effectively improves the performance of self-supervised learning and achieves state-of-the-art performance on different downstream tasks.
For example, it achieves state-of-the-art 75.7\% ImageNet top-1 accuracy under the pre-training setting of 1024 batch size and 800 epochs for ResNet50.

The main contributions of this study are threefold:
\begin{itemize}
\item A novel contrastive learning method, i.e., \method, is proposed, and it can enhance semantic alignment in contrastive learning.
To our knowledge, this is the first work that defines a feature search task in contrastive learning.
\item We expand contrastive learning from a data-to-data manner to a feature-to-data manner, which enriches the diversity of augmentations.
\item The proposed \method achieves state-of-the-art performance on different downstream tasks.
\end{itemize}
\section{Related Works}
\label{sec:relatedwork}

Recently, some studies \cite{scrl_2021,resim_2021,soco_2021,detco_2021,ORL_2021,univip_2022,pixpro_2021,densecl_2021,dsc_2021,setsim_2021} pointed out that the problem of semantic inconsistency is more serious for downstream dense prediction tasks, such as object detection and instance segmentation.
Therefore, these methods utilize region-level and pixel-level features for contrast.
In this study, the proposed \method construct feature-level augmentations using dense feature maps.
Therefore, this section introduces related studies that conduct contrastive learning using region-level and pixel-level features.

\textbf{Region-level contrastive learning.}
SCRL \cite{scrl_2021} minimizes the distance between two local features, which are cropped from two corresponding feature maps of two views.
ReSim \cite{resim_2021} aligns regional representations by sliding a fixed-sized window across the overlapping area between two views to improve the performance for localization-based tasks.
SoCo \cite{soco_2021}, ORL \cite{ORL_2021}, and UniVIP \cite{univip_2022} extract object region proposals and use them to construct region-level features for contrastive learning. 
They achieve good performance for downstream dense prediction tasks.

\textbf{Pixel-level contrastive learning.}
To obtain a more fine-grained representation, several studies \cite{pixpro_2021,densecl_2021,setsim_2021} design pixel-level contrastive learning task, which assumes that features extracted from the same pixel of different views should be treated as positive pairs while pixels from others must be distinguished.
PixPro \cite{pixpro_2021} utilizes a pixel propagation module to select similar pixel features for contrast and encourages consistency between positive pixel pairs.
DenseCL \cite{densecl_2021} proposes a dense projection head to generate dense feature vectors for pixel-level contrastive learning.
SetSim \cite{setsim_2021} is designed to realize pixel-wise similarity learning by filtering out noisy backgrounds.

As summarized above, data augmentations bring rich information while increasing uncertainty in contrastive learning.
While methods that utilize region-level features expand the granularity of feature representation by alleviating the influence of noises.
Unlike previous studies, our method bridges the correlation between data and feature augmentations and extends the contrastive-based self-supervised task to a semantics-consistent feature search task.
\section{Methods}
\label{sec:methods}

In this section, we first introduce the overall architecture of \method in \cref{subsec_overall}.
Then, the contrast between data augmentations is presented in \cref{subsec_ldd}.
Next, the key feature search module of \method is introduced in detail in \cref{subsec_SCFS}.
Finally, the implementation details are presented in \cref{subsec_implementation}.

\begin{figure*}[!tp]
\centering
\includegraphics[height=5cm]{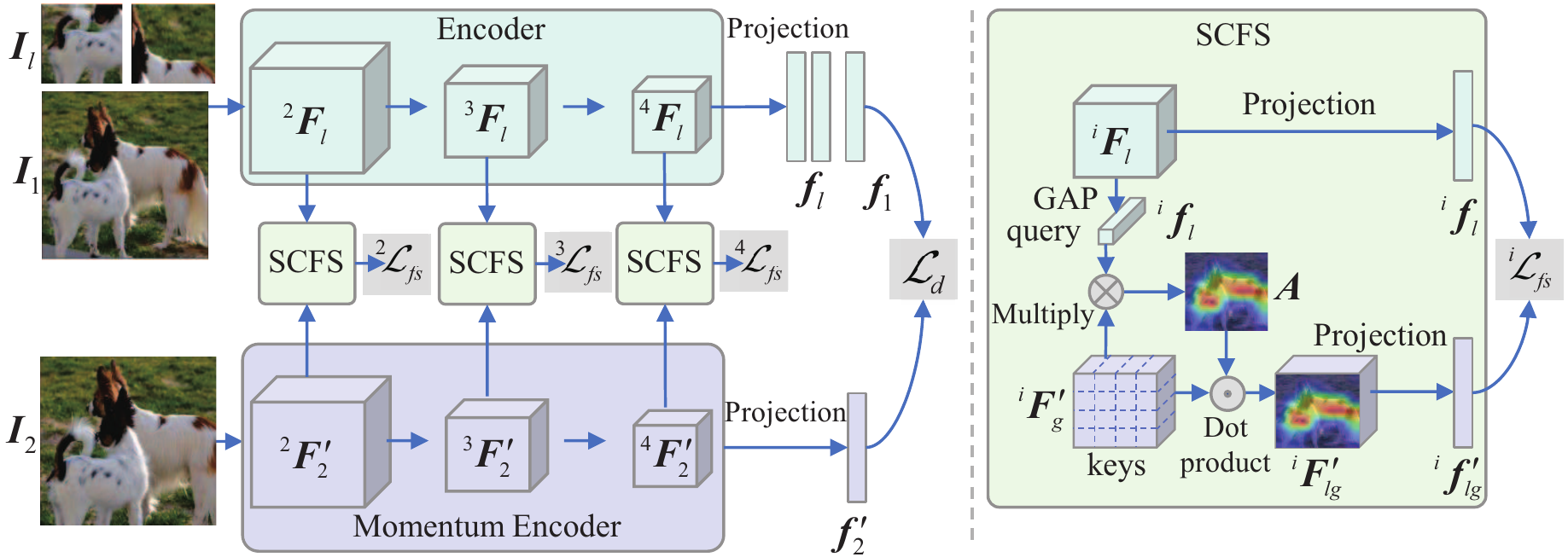}
\caption{
Overall architecture of the proposed semantics-consistent feature search (\method).
It consists of an encoder and a momentum encoder.
There are two contrastive learning tasks: the contrast between data augmentations ($\mathcal{L}_{d}$) in the final feature space and the feature search task conducted on multiple layers ($\mathcal{L}_{fs}$).
The details of the feature search procedure is shown on the right.
}
\label{fig:method}
\end{figure*}

\subsection{Overall Architecture}
\label{subsec_overall}

The overall architecture of \method is shown in \cref{fig:method}.
It consists of an encoder and a momentum encoder.
The momentum encoder is an exponential-moving-average version of the encoder.
\method consists of two contrastive learning tasks: the contrast between data augmentations ($\mathcal{L}_{d}$) and the contrast between data augmentations and feature augmentations ($\mathcal{L}_{fs}$).

\textbf{The contrast between data augmentations.}
Given two global augmentations ($\vm{I}_1$ and $\vm{I}_2$) and multiple local augmentations $\vm{I}_l$ of an input image, the final output feature representations $\vm{f}$ of data augmentations are utilized to calculate the contrastive loss $\mathcal{L}_{d}$ (which will be introduced in the second subsection).

\textbf{The contrast between data augmentations and feature augmentations.}
As introduced in \cref{sec:intro}, it is unavoidable to construct augmentations that contain different semantic concepts during the augmentation procedure.
It's harmful to pull these augmentations close indiscriminately in the feature space.
Therefore, we propose the \method (which will be introduced in the third subsection) method to enhance the contrast between semantics-consistent regions in different augmentations.
As shown in \cref{fig:method}, to fully enhance the contrast between semantics-consistent features, \method is employed on multiple layers of the backbone network.
At the $i$-th layer, \method utilizes the feature ${}^i{\vm{f}_l}$ from the encoder to search semantics-consistent feature ${}^i{\vm{f'}_{lg}}$ on the feature map ${}^i{\vm{F'}_g}$ from the momentum encoder.
And a feature search loss ${}^i\mathcal{L}_{fs}$ is calculated between the data augmentation ${}^i{\vm{f}_l}$ and the feature augmentation ${}^i{\vm{f'}_{lg}}$.
The overall feature search loss is the sum of all layers:
\begin{equation}
\mathcal{L}_{fs} = \sum\limits_{i \in {V_L}} {{}^i\mathcal{L}_{fs}}
\label{equ:l_fs}
\end{equation}
where $V_L$ denotes the set of layers to conduct \method.

The overall loss is the sum of the contrastive loss between data augmentations and the feature search loss:
\begin{equation}
\mathcal{L} = {\mathcal{L}_d} + \mathcal{L}_{fs}
\label{equ:l_SCFS}
\end{equation}

\subsection{Contrast Between Data Augmentations}
\label{subsec_ldd}

Given a pair of global augmentations ($\vm{I}_1$ and $\vm{I}_2$) of an input image, the feature representations of the two augmentations are used to calculate the global contrastive loss.
Specifically, ${\vm{f}_{1}} = {E_{\vm{\theta}} }\left( {{\vm{I}_{1}}} \right)$ and ${\vm{f'}_{2}} = {E_{\vm{\theta}'} }\left( {{\vm{I}_2}} \right)$, where $\vm{\theta}$ and $\vm{\theta}'$ are parameters of the encoder and the momentum encoder, respectively.
${\vm{f}_1}$, ${\vm{f'}_2} \in {R^K}$, $K$ is the output dimension.
$\vm{f}_1$ is normalized with a softmax function:
\begin{equation}
P_1^i = \frac{{\exp \left( {{{f_1^i} \mathord{\left/
 {\vphantom {{f_1^i} \tau }} \right.
 \kern-\nulldelimiterspace} \tau }} \right)}}{{\sum\nolimits_{k = 1}^K {\exp \left( {{{f_1^k} \mathord{\left/
 {\vphantom {{f_1^k} \tau }} \right.
 \kern-\nulldelimiterspace} \tau }} \right)} }}
\label{equ:l_softmax}
\end{equation}

where $\tau>0$ is a temperature parameter that controls the sharpness of the output distribution.
Note that ${P'_2}$ is obtained by normalizing ${\vm{f'}_2}$ with a similar softmax function with temperature $\tau'$.
$\vm{I}_1$ and $\vm{I}_2$ are fed to the momentum encoder and encoder symmetrically, and ${P'_1}$ and ${P_2}$ are obtained respectively.
Following DINO \cite{dino_2021}, the cross-entropy loss is employed as the contrastive loss between two global views:
\begin{equation}
{\mathcal{L}_{g}} =  - \left( {P'_2}\log \left( {{P_1}} \right) + {P'_1}\log \left( {{P_2}} \right) \right)
\label{equ:l_gg}
\end{equation}

To enrich augmentations, the multi-crop strategy \cite{swav_2020} is employed.
Multiple local augmentations $\vm{I}_l$ is also constructed and fed to the encoder: ${\vm{f}_l}\!=\!{E_{\vm{\theta}}}\left( {{\vm{I}_l}} \right)$.
${P_l}$ is obtained by normalizing ${\vm{f}_l}$ with the softmax function with temperature $\tau$.
The contrast between local views and global views can be calculated:
\begin{equation}
{\mathcal{L}_{l}} = \sum\limits_{n = 1}^N { - \left( {{{P'}_1}\log \left( {P_l^n} \right) + {{P'}_2}\log \left( {P_l^n} \right)} \right)}
\label{equ:l_lgd}
\end{equation}
where $N$ denotes the number of local views.
Thus, the overall loss is the sum of global loss and local loss:
\begin{equation}
{\mathcal{L}_{d}} = {\mathcal{L}_g} + \mathcal{L}_{l}
\label{equ:l_mc}
\end{equation}

\subsection{Semantics-Consistent Feature Search}
\label{subsec_SCFS}

We propose \method to enhance the importance of semantics-consistent regions in different augmentations by conducting contrast learning between data and feature augmentations.

The architecture of \method is shown in \cref{fig:method}.
By feeding the local augmentations $\vm{I}_{l}$ to the encoder, feature maps from different stages of the backbone ResNet50 \cite{resnet_2016} are extracted.
Specifically, the output features from different stages, i.e., $Res2$, $Res3$ and $Res4$, are utilized to conduct \method, ensuring that each stage of the backbone produces discriminative features:
$\left\{ {{}^2{\vm{F}_l},{}^3{\vm{F}_l},{}^4{\vm{F}_l}} \right\} = {E_{\vm{\theta}} }\left( {{\vm{I}_{l}}} \right)$, where ${}^i{\vm{F}_l} \in {R^{{W_l^i} \times {H_l^i} \times {C^i}}}$, ${W_l^i},{H_l^i},C^i$ denote the width, height and channel dimension, respectively.
Next, the global average pooling operation is conducted on each ${}^i{\vm{F}_l}$ in the spatial dimensions:
\begin{equation}
{}^i{\vm{f}_l} = \frac{1}{{{W_l^i} \times {H_l^i}}}\sum\limits_{x = 1}^{{W_l^i}} {\sum\limits_{y = 1}^{{H_l^i}} {{}^i{\vm{F}_l}\left( {x,y,z} \right)} }
\label{equ:gap}
\end{equation}
where ${}^i{\vm{f}_l} \in {R^{C^i}}$.
Meanwhile, the global augmentations $\vm{I}_g$ ($g=1,2$) are fed to the momentum encoder to extract feature maps from different stages: $\left\{ {{}^2{\vm{F}'_g},{}^3{\vm{F}'_g},{}^4{\vm{F}'_g}} \right\} = {E_{\vm{\theta}'} }\left( {{\vm{I}_{g}}} \right)$, ${}^i{\vm{F}'_g} \in {R^{{W_g^i} \times {H_g^i} \times {C^i}}}$, ${W_g^i},{H_g^i},C^i$ denote width, height and channel dimension, respectively.

Then, based on ${}^i{\vm{f}_l}$ and ${}^i{\vm{F}'_g}$, \method aims to adaptively search the most semantics-consistent features in ${}^i{\vm{F}'_g}$ for contrast, while suppressing irrelevant features.
In \method, each feature ${}^i{\vm{f}_l}$ of the local data augmentations is treated as query, and the features ${}^i{\vm{F}'_g}$ of the global augmentations are treated as keys.
The similarity between ${}^i{\vm{f}_l}$ and ${}^i{\vm{F}'_g}$ is calculated:
\begin{equation}
\vm{A}\left( {x,y} \right) = \frac{{{}^i{\vm{f}_l} \cdot {{}^i{\vm{F}'_g}}\left( {x,y} \right)}}{{{{\left\| {{}^i{\vm{f}_l}} \right\|}_2}{{\left\| {{{}^i{\vm{F}'_g}}\left( {x,y} \right)} \right\|}_2}}}
\label{equ:attention}
\end{equation}
where $\vm{A} \in {R^{{W_g^i} \times {H_g^i}}}$ is the attention map, and $x = 1, \ldots ,{W_g^i}$, $y = 1, \ldots ,{H_g^i}$, ${\left\|  \cdot  \right\|_2}$ is the L2 norm.
The attention map $\vm{A}$ activates the semantics-consistent regions of the local augmentation on the global augmentation.
Thus, the higher portion of local regions can be searched.
To select semantic features and suppress irrelevant local features, we directly multiply the attention map $\vm{A}$ with ${}^i{\vm{F}'_g}$ to obtain the semantics-consistent feature augmentations:
\begin{equation}
{}^i{\vm{F}'_{lg}} = \vm{A} \cdot {}^i{\vm{F}'_g}
\label{equ:attention}
\end{equation}
This operation can be regarded as attention-weighted average pooling.
Through feature search, $N$ local data augmentations $\vm{I}_{l}$ can search $N$ corresponding semantics-consistent features ${}^i{\vm{F}'_{lg}}$ from a global data augmentation $\vm{I}_{g}$.
That is, in terms of the global data augmentation, $N$ different features are constructed in the feature space through the feature search procedure.
Therefore, we term the searched semantics-consistent features ${}^i{\vm{F}'_{lg}}$ as feature-level augmentations.
After \method, the feature augmentation ${}^i{\vm{F}'_{lg}}$ only contains region-level features which are semantic-related to the local augmentation ${}^i{\vm{F}_l}$.

Next, ${}^i{\vm{F}_l}$ and ${}^i{\vm{F}'_{lg}}$ are fed to corresponding projection heads to obtain their final representations for contrast:
\begin{equation}
\left\{ \begin{array}{l}
{}^i{\vm{f}_l} = {{\mathop{\rm H}\nolimits} _i}\left( {{}^i{\vm{F}_l}} \right) \\
{}^i{\vm{f}'_{lg}} = {{H'}_i}\left( {{}^i{\vm{F}'_{lg}}} \right)
\end{array} \right.
\label{equ:head}
\end{equation}
where ${{\mathop{\rm H}\nolimits} _i}$ and ${\rm{H}'}_i$ denote the projection heads on the $i$-th layer of the encoder and the momentum encoder, respectively.
${}^i{\vm{f}_l}$ and ${}^i{\vm{f}'_{lg}}$ are normalized with softmax function with temperature $\tau$ and $\tau'$, respectively, as the same formulation in \cref{equ:l_softmax}.
The corresponding output probability ${}^i{P_l}$ and ${}^i{P'_{lg}}$ are employed to calculate the contrast loss between local data augmentations and feature augmentations:
\begin{equation}
{}^i{\mathcal{L}_{fs}} = \sum\limits_{g = 1}^2 {\sum\limits_{n = 1}^N { - \left( {{}^i{{P'}_{lg}}\log \left( {{}^iP_l^n} \right)} \right)} }
\label{equ:l_lldfi}
\end{equation}

Through \method, the contrast between feature augmentations and data augmentations is bridged.
The model can adaptively search the semantics-consistent features for contrast.
Therefore, it can enhance the importance of semantics-consistent regions in different augmentations, alleviating the uncertainty in contrastive learning introduced by data augmentations that contain different semantic concepts.

\subsection{Implementation Details}
\label{subsec_implementation}

\method is based on DINO \cite{dino_2021} and we follow the most hyper-parameter settings of DINO.
For a fair comparison, the standard ResNet50 \cite{resnet_2016} is employed as the backbone network in all experiments.

For data augmentation, the global augmentations consist of random cropping, resizing to $224 \times 224$, random horizontal flip, gaussian blur, and color jittering.
And the local augmentations consist of random cropping, resizing to $96 \times 96$, random horizontal flip, gaussian blur, and color jittering.
For feature augmentations in \method, the $Res2$, $Res3$, and $Res4$ layers are used.
Two global views with $N=8$ local views are the default setting of augmentation.

The projection head for the contrast between data augmentations consists of a four-layer multi-layer-perceptron (MLP) with the same architecture as DINO \cite{dino_2021}.
The projection head for feature search consists of three convolutional layers and two FC layers.

The Pytorch-style pseudocode of \method is shown in Algorithm \ref{algo}.
For simplification, we only show one local augmentation and the $i$-th layer for feature search.

\begin{algorithm}[]
\caption{PyTorch-style pseudocode of \method.}
\label{algo}
\begin{lstlisting}[language=python]
# es, et: encoder and momentum encoder networks
# hs_i, ht_i: head on the layer-i for feature search of the encoder and momentum encoder
# C, Ci: centers
# tps, tpt: temperatures
# l, m: network and center momentum rates
et.params = es.params
for I in loader: # load a minibatch I with n samples
    I1, I2 = augment(I), augment(I) # global views
    Il = augment(I) # multiple local views
    # encoder output
    s1, _, = es(I1)
    s2, _, = es(I2)
    sl, Sl_i = es(Il)
    # momentum encoder output
    t1, T1_i = es(I1)
    t2, T2_i = es(I2)
    # feature search
    sl_i_1, t1_i = FS(Sl_i, T1_i, hs_i, ht_i)
    sl_i_2, t2_i = FS(Sl_i, T2_i, hs_i, ht_i)
    # contrastive loss for data augmentation
    loss_g = H(t1, s2, C)/2 + H(t2, s1, C)/2
    loss_l = H(t1, sl, C)/2 + H(t2, sl, C)/2
    loss_d = loss_g + loss_l
    # feature search loss
    loss_fs = H(t1_i, sl_i_1, Ci)/2 + H(t2_i, sl_i_2, Ci)/2
    # total loss
    loss = loss_d + loss_fs
    loss.backward() # back-propagate
    # encoder, momentum encoder and center updates
    update(es) # SGD
    et.params = l*et.params + (1-l)*es.params
    C = m*C + (1-m)*cat([t1, t2]).mean(dim=0)
    Ci = m*Ci + (1-m)*cat([t1_i, t2_i]).mean(dim=0)

def H(t, s, C):
    t = t.detach() # stop gradient
    s = softmax(s / tps, dim=1)
    t = softmax((t - C) / tpt, dim=1) # center + sharpen
    return - (t * log(s)).sum(dim=1).mean()

def FS(t, s, hs, ht):
    t = t.detach() # stop gradient
    s = gap(s, dim=(1,2)) # gap
    s = normalize(s, dim=1) # l2-normalize
    t = normalize(t, dim=3) # l2-normalize
    a = (s*t).sum(dim=3) # similarity
    s = a*s
    return hs(s), ht(t)
\end{lstlisting}
\end{algorithm}

\section{Experiments}
\label{sec:experiments}

In this section, comprehensive experiments are conducted to demonstrate the effectiveness of \method.
We evaluate the performance on different downstream tasks, including ImageNet classification, object detection, instance segmentation, and other classification task on small datasets.
In addition, we conduct ablation experiments to analyze the influence of each component in \method.

\setlength{\tabcolsep}{4pt}
\begin{table}[!t]
\begin{center}
\begin{tabular}{lllccc}
\toprule
\noalign{\smallskip}
Method & Batch Size & Epochs & LP & $k$-NN \\
\noalign{\smallskip}
\hline
\noalign{\smallskip}
Supervised  & 256 & 100 & 76.2 & 74.8 \\
SimCLR \cite{simclr_2020} & 4096 & 1000 & 69.3 & - \\
BYOL \cite{byol_2020} & 4096 & 1000 & 74.3 & 66.9 \\
BYOL \cite{byol_2020} & 4096 & 200 & 70.6 & - \\
SwAV \cite{swav_2020} & 4096 & 800 & 75.3 & - \\
SwAV \cite{swav_2020} & 256 & 200 & 72.7 & - \\
MoCo-v2 \cite{moco_2020} & 256 & 200 & 67.5 & 54.3 \\
SimSiam \cite{simsiam_2021} & 256 & 200 & 70.0 & - \\
ISD \cite{isd_2021} & 256 & 200 & 69.8 & 62.0 \\
MSF \cite{msf_2021} & 256 & 200 & 71.4 & 64.0 \\
NNCLR \cite{nnclr_2021} & 4096 & 200 & 70.7 & - \\
Barlow Twins \cite{barlowtwins_2021} & 2048 & 1000 & 73.2 & - \\
VICReg \cite{vicreg_2021} & 2048 & 1000 & 73.2 & - \\
OBoW \cite{obow_2021} & 256 & 200 & 73.8 & - \\
DCL \cite{dcl_2021} & 256 & 200 & 66.9 & - \\
CLSA \cite{clsa_2021} & 256 & 200 & 73.3 & - \\
AdCo \cite{adco_2021} & 256 & 200 & 73.2 & - \\
DetCo \cite{detco_2021} & 256 & 200 & 68.6 & - \\
UniVIP \cite{univip_2022} & 4096 & 200 & 73.1 & - \\
HCSC \cite{hcsc_2022} & 256 & 200 & 73.3 & - \\
MoCo-v3 \cite{mocov3_2021} & 4096 & 300 & 72.8 & - \\
MoCo-v3 \cite{mocov3_2021} & 4096 & 1000 & 74.6 & - \\
DINO* \cite{dino_2021} & 256 & 200 & 73.0 & 64.0 \\
DINO \cite{dino_2021} & 4080 & 800 & 75.3 & 67.5 \\
\rowcolor{Gray} \textbf{\method} & 256 & 200 & \underline{73.9} & \underline{65.5} \\
\rowcolor{Gray}\textbf{\method} & 1024 & 800 & \textbf{75.7} & \textbf{68.5} \\
\bottomrule
\end{tabular}
\end{center}
\setlength{\abovecaptionskip}{0.05cm}
\caption{Linear probing and $k$-NN accuracy (\%) on ImageNet. The result with ``*" is reproduced for fair comparison. LP denotes linear probing. Bold font and underline indicate the best results under the setting of 256 batch size and 200 epochs and the setting of 1024 batch size and 800 epochs, respectively.}
\label{table:knnlinear}
\end{table}

\begin{table}[!t]
\begin{center}
\setlength{\tabcolsep}{0.45mm}{
\begin{tabular}{lllcccc}
\toprule
\noalign{\smallskip}
\multirow{2}*{Method} & \multirow{2}*{Batch Size} & \multirow{2}*{Epochs} & \multicolumn{2}{c}{Top-1} & \multicolumn{2}{c}{Top-5} \\
~ & ~ & ~ & 1\% & 10\% & 1\% & 10\% \\
\noalign{\smallskip}
\hline
\noalign{\smallskip}
Supervised \cite{s4l_2019} & 256 & 90 & 25.4 & 56.4 & 48.4 & 80.4 \\
SimCLR \cite{simclr_2020} & 4096 & 1000 & 48.3 & 65.6 & 75.5 & 87.8 \\
BYOL \cite{byol_2020} & 4096 & 1000 & 53.2 & 68.8 & 78.4 & 89.0 \\
SwAV \cite{swav_2020} & 4096 & 800 & 53.9 & 70.2 & 78.5 & 89.9 \\
DINO \cite{dino_2021} & 4080 & 800 & 50.2 & 69.3 & 74.0 & 89.1 \\
\rowcolor{Gray}\textbf{\method} & 1024 & 800 & \textbf{54.3} & \textbf{70.5} & \textbf{78.6} & \textbf{90.2} \\
\bottomrule
\end{tabular}}
\end{center}
\setlength{\abovecaptionskip}{0.05cm}
\caption{Evaluation on small labeled ImageNet. Bold font indicates the best result.}
\label{table:semisupervised}
\end{table}

\subsection{Comparing with SSL methods on ImageNet}

\textbf{$k$-NN and Linear Probing Accuracy on ImageNet.}
After pre-training on the ImageNet ILSVRC-2012 \cite{imagenet_2015} training set, the pre-trained models are evaluated on the ImageNet ILSVRC-2012 validation set.
For $k$-NN, it is evaluated as in study \cite{insdis_2018}.
For linear probing, we train a linear classifier from scratch based on the feature extracted by a fixed backbone with 100 epochs \cite{moco_2020}.
The top-1 accuracy is adopted as the evaluation metric.

The results are reported in Table \ref{table:knnlinear}.
With the standard ResNet50 \cite{resnet_2016} architecture and pre-trained with 256 batch size for 200 epoch, the proposed \method achieves the best $k$-NN top-1 accuracy 65.5\% and the best linear probing top-1 accuracy 73.9\%, outperforming its baseline DINO \cite{dino_2021} by 1.5\% and 0.9\%, respectively.
In addition, with 1024 batch size and 800 epoch, \method achieves the best $k$-NN accuracy (68.5\%) and linear probing accuracy (75.7\%), outperforming the accuracy of DINO \cite{dino_2021} trained with 4080 batch size for 800 epoch.
This result demonstrates that \method can improve the representation learning performance by searching semantics-consistent features for contrast.

\setlength{\tabcolsep}{8pt}
\begin{table}[!t]
\begin{center}
\begin{tabular}{llccc}
\toprule
Method & Epochs & $\rm{AP}^{\rm{b}}$ & $\rm{AP}_{50}^{\rm{b}}$ & $\rm{AP}_{75}^{\rm{b}}$ \\
\hline
Scratch & - & 33.8 & 60.2 & 33.1 \\
Supervised & 90 & 53.5 & 81.3 & 58.8 \\
SimCLR \cite{simclr_2020} & 1000 & 56.3 & 81.9 & 62.5 \\
BYOL \cite{byol_2020} & 300 & 51.9 & 81.0 & 56.5 \\
SwAV \cite{swav_2020} & 400 & 45.1 & 77.4 & 46.5 \\
DINO \cite{dino_2021} & 800 & 55.9 & 82.1 & 62.3 \\
\rowcolor{Gray}\textbf{\method} & 800 & \textbf{57.4} & \textbf{83.0} & \textbf{63.6} \\
\bottomrule
\end{tabular}
\end{center}
\setlength{\abovecaptionskip}{0.05cm}
\caption{Results for PASCAL VOC object detection using Faster R-CNN \cite{fasterrcnn_2015} with ResNet50-C4. Bold font indicates the best result.}
\label{table:voc}
\end{table}

\setlength{\tabcolsep}{0.9mm}
\begin{table*}[t]
\centering
\begin{tabular}{l|c|cccccc|cccccc}
\toprule
\multirow{2}{*}{Method} & \multirow{2}{*}{Epochs} & \multicolumn{6}{c}{$1\!\times\!\rm{schedule}$} \vline & \multicolumn{6}{c}{$2\!\times\!\rm{schedule}$} \\
 &  & $\rm{AP}^{\rm{b}}$ & $\rm{AP}_{50}^{\rm{b}}$ & $\rm{AP}_{75}^{\rm{b}}$ & $\rm{AP}^{\rm{s}}$ & $\rm{AP}_{50}^{\rm{s}}$ & $\rm{AP}_{75}^{\rm{s}}$ & $\rm{AP}^{\rm{b}}$ & $\rm{AP}_{50}^{\rm{b}}$ & $\rm{AP}_{75}^{\rm{b}}$ & $\rm{AP}^{\rm{s}}$ & $\rm{AP}_{50}^{\rm{s}}$ & $\rm{AP}_{75}^{\rm{s}}$ \\ \hline
Scratch & - & 31.0 & 49.5 & 33.2 & 28.5 & 46.8 & 30.4 & 38.4 & 57.5 & 42.0 & 34.7 & 54.8 & 37.2  \\
Supervised & 90 & 38.9 & 59.6 & 42.7 & 35.4 & 56.5 & 38.1 & 41.3 & 61.3 & 45.0 & 37.3 & 58.3 & 40.3 \\
\midrule
MoCo~\cite{moco_2020} & 200 & 38.5 & 58.9 & 42.0 & 35.1 & 55.9 & 37.7 & 40.8 & 61.6 & 44.7 & 36.9 & 58.4 & 39.7 \\
MoCo v2~\cite{mocov2_2020} & 200 & 40.4 & 60.2 & 44.2 & 36.4 & 57.2 & 38.9 & 41.7 & 61.6 & 45.6 & 37.6 & 58.7 & 40.5 \\
BYOL~\cite{byol_2020} & 300 & 40.4 & 61.6 & 44.1 & \textbf{37.2} & 58.8 & \textbf{39.8} & 42.3 & 62.6 & 46.2 & \textbf{38.3} & 59.6 & \textbf{41.1} \\
SwAV~\cite{swav_2020} & 400 & - & - & - & - & - & - & 42.3 & 62.8 & \textbf{46.3} & 38.2 & 60.0 & 41.0 \\
ReSim-FPN$^{T}$~\cite{resim_2021} & 200 & 39.8 & 60.2 & 43.5 & 36.0 & 57.1 & 38.6 & 41.4 & 61.9 & 45.4 & 37.5 & 59.1 & 40.3 \\
SetSim~\cite{setsim_2021} & 200 & 40.2 & 60.7 & 43.9 & 36.4 & 57.7 & 39.0 & 41.6 & 62.4 & 45.9 & 37.7 & 59.4 & 40.6 \\
DenseCL~\cite{densecl_2021} & 200 & 40.3 & 59.9 & 44.3 & 36.4 & 57.0 & 39.2 & 41.2 & 61.9 & 45.1 & 37.3 & 58.9 & 40.1 \\
DSC~\cite{dsc_2021} & 200 & 39.4 & 58.9 & 43.2 & 35.7 & 56.1 & 38.3 & - & - & - & - & - & - \\
HSA~\cite{hsa_2022} & 800 & 40.2 & 60.9 & 43.9 & 36.5 & 57.9 & 39.1 & \textbf{42.2} & 63.0 & 46.1 & 38.1 & 59.9 & 40.9 \\
DetCo~\cite{detco_2021} & 800 & 40.1 & 61.0 & 43.9 & 36.4 & 58.0 & 38.9 & - & - & - & - & - & - \\
ORL*~\cite{ORL_2021} & 800 & 40.3 & 60.2 & \textbf{44.4} & 36.3 & 57.3 & 38.9 & - & - & - & - & - & - \\
DINO~\cite{dino_2021} & 800 & 40.0 & 61.6 & 43.4 & 36.5 & 58.6 & 39.1 & 41.9 & 62.6 & 46.0 & 37.8 & 59.7 & 40.6 \\
\rowcolor{Gray}\textbf{\method} & 800 & \textbf{40.5} & \textbf{61.8} & 44.0 & 36.7 & \textbf{58.8} & 39.2 & 42.1 & \textbf{63.4} & 46.1 & 38.1 & \textbf{60.2} & 41.0 \\
\bottomrule
\end{tabular}
\caption{Object detection and instance segmentation on COCO using Mask R-CNN \cite{maskrcnn_2017} with ResNet50-FPN. Bold font indicates the best result.}
\label{tab_coco}
\end{table*}

\noindent
\textbf{Semi-Supervised Learning on ImageNet.}
In this part, we evaluate the performance of \method under the semi-supervised setting.
Specifically, we use 1\% and 10\% of the labeled training data from ImageNet \cite{imagenet_2015} for finetuning, which follows the semi-supervised protocol in SimCLR \cite{simclr_2020}.
The same splits of 1\% and 10\% of ImageNet labeled training data in SimCLRv2 \cite{simclrv2_2020} are used.

The results are reported in Table \ref{table:semisupervised}.
After finetuning using 1\% and 10\% training data, \method outperforms all the compared methods.
The results demonstrate that \method achieves the best feature representation quality.

\setlength{\tabcolsep}{1.2mm}
\begin{table*}[t]
\begin{center}
\begin{tabular}{lccccccc}
\toprule
Method & CIFAR-10 & CIFAR-100 & CUB-Bird & Stanford-Cars & Aircraft  & Oxford-Pets \\ \hline
Supervised &97.5&86.4&81.3&92.1&86.0&92.1\\
SimCLR\cite{simclr_2020}  &97.7 & 85.9 & --& 91.3& 88.1 &89.2\\
BYOL\cite{byol_2020}  & 97.8 & 86.1 & -- & 91.6 & 88.1 & 91.7 \\
DINO\cite{dino_2021}*  & 97.7 & 86.6 & 81.0 &91.1 & 87.4 & 91.5\\
\rowcolor{Gray}\textbf{\method}  & \textbf{97.8} & \textbf{86.7} & \textbf{82.7} & \textbf{91.6} & \textbf{88.5} & \textbf{91.9} \\
\bottomrule
\end{tabular}
\end{center}
\setlength{\abovecaptionskip}{0.05cm}
\caption{Transfer learning results from ImageNet with the standard ResNet50 \cite{resnet_2016}. * denotes the results are reproduced in this study. Bold font indicates the best result.}
\label{table:Transfer learning}
\end{table*}

\subsection{Transfer Learning on Downstream Tasks}

\textbf{Object Detection and Instance Segmentation.}
In this part, we evaluate the representations of \method on dense prediction tasks, i.e., object detection and instance segmentation, on mainstream datasets PASCAL VOC \cite{voc_2010} and MS COCO \cite{coco_2014} datasets.
On the PASCAL VOC dataset \cite{voc_2010}, the trainval07+12 set is used as the training set, and the test2007 set is used as the test set.
Following \cite{soco_2021}, Faster R-CNN detector \cite{fasterrcnn_2015} with the ResNet50-C4 backbone initialized by the self-supervised pre-trained model is trained end-to-end.
On the COCO dataset, the train2017 set is used for training and the val2017 set is used for evaluation.
The Mask R-CNN \cite{maskrcnn_2017} with R50-FPN is used.
The $\rm{AP}^{\rm{b}}$, $\rm{AP}_{50}^{\rm{b}}$ and $\rm{AP}_{75}^{\rm{b}}$ metrics are used for object detection.
While the $\rm{AP}^{\rm{s}}$, $\rm{AP}_{50}^{\rm{s}}$ and $\rm{AP}_{75}^{\rm{s}}$ metrics are used for instance segmentation.

The experimental results are shown in Table \ref{table:voc} and Table \ref{tab_coco}.
\method achieves best performance on the two datasets.
For example, on VOC, \method achieves 57.4\% $\rm{AP}^{\rm{b}}$, 83.0\% $\rm{AP}_{50}^{\rm{b}}$ and 63.6\% $\rm{AP}_{75}^{\rm{b}}$.
The $\rm{AP}^{\rm{b}}$ of \method outperforms its baseline DINO by 1.5\%.
These results shows that \method also has good transfer ability on dense prediction tasks.

\noindent
\textbf{Other Classification Tasks.}
In this part, we focus on the performance of self-supervised models when they are finetuned on small datasets, including CIFAR \cite{cifar10} and fine grained datasets \cite{car,cub,aircraft,pets}.
The results are shown in Table \ref{table:Transfer learning}.
The proposed \method shows the best performance on all the small datasets, which demonstrates that \method has good generalization ability.

\subsection{Pre-training on Uncured Dataset}
The proposed \method can solve the problem of semantic inconsistency during pre-training, which is important when pre-training on uncured datasets since this problem is more serious.
To verify this, we pre-train \method and DINO on COCO \cite{coco_2014}, which is much more uncured than ImageNet.
The same hyper-parameters used on ImageNet are applied to train the models with 512 batch size for 500 epochs.
After pre-training, we fine-tune the pre-trained models on COCO for object detection and instance segmentation.
The Mask R-CNN \cite{maskrcnn_2017} with R50-FPN is used.
As shown in Table \ref{tab_pt_on_coco}, \method improves the performance significantly compared to its baseline DINO.
In addition, when compared to other dense pixel-level and region-level methods, such as DenseCL \cite{densecl_2021} and ORL \cite{ORL_2021}, \method also achieves the best performance.
This experiment verifies that \method can effectively solve the problem of semantic inconsistency during pre-training.

\setlength{\tabcolsep}{0.4mm}
\begin{table}[t]
\centering
\begin{tabular}{l|c|cccccc}
\toprule
Method & Pre-train & $\rm{AP}^{\rm{b}}$ & $\rm{AP}_{50}^{\rm{b}}$ & $\rm{AP}_{75}^{\rm{b}}$ & $\rm{AP}^{\rm{s}}$ & $\rm{AP}_{50}^{\rm{s}}$ & $\rm{AP}_{75}^{\rm{s}}$ \\ \hline
Scratch & - & 31.0 & 49.5 & 33.2 & 28.5 & 46.8 & 30.4 \\
Supervised & ImageNet & 38.9 & 59.6 & 42.7 & 35.4 & 56.5 & 38.1 \\
SimCLR\cite{simclr_2020} & COCO & 37.0 & 56.8 & 40.3 & 33.7 & 53.8 & 36.1 \\
MoCov2\cite{mocov2_2020} & COCO & 38.5 & 58.1 & 42.1 & 34.8 & 55.3 & 37.3 \\
BYOL\cite{byol_2020} & COCO & 39.5 & 59.3 & 43.2 & 35.6 & 56.5 & 38.2 \\
DenseCL\cite{densecl_2021} & COCO & 39.6 & 59.3 & 43.3 & 35.7 & 56.5 & 38.4 \\
ORL\cite{ORL_2021} & COCO & 40.3 & 60.2 & 44.4 & 36.3 & 57.3 & 38.9 \\
UniVIP\cite{univip_2022} & COCO & 40.8 & - & - & 36.8 & - & - \\
DINO\cite{dino_2021} & COCO & 39.0 & 59.6 & 42.9 & 35.6 & 56.8 & 38.0 \\
\rowcolor{Gray}\textbf{\method} & COCO & \textbf{40.9} & \textbf{61.6} & \textbf{44.4} & \textbf{36.9} & \textbf{58.4} & \textbf{39.5} \\
\bottomrule
\end{tabular}
\caption{Pre-training and than Fine-tuning on COCO using Mask R-CNN \cite{maskrcnn_2017} with ResNet50-FPN and 1$\times$ schedule. All models pre-trained on COCO are pre-trained with 512 batch size for 800 epochs. Bold font indicates the best result.}
\label{tab_pt_on_coco}
\end{table}

\subsection{Ablation Studies}
We analyze the influence of each component in \method.
To speed up the training time, the ImageNet100 dataset, which contains 100 randomly selected categories from ImageNet \cite{imagenet_2015}, is adopted.
All the models are pre-trained on the ImageNet100 training set with 256 batch size for 200 epoch, and tested on the validation set.
The $k$-NN and linear probing top-1 accuracy are used as the evaluation metrics.

\noindent
\textbf{Influence of Different Contrast Modes.}
The contrast mode can be divided into three types:
contrast between two global data augmentations used in all contrastive learning methods ($\rm{G_d2G_d}$);
contrast between local data augmentations and global data augmentations used in multi-crop strategy ($\rm{L_d2G_d}$);
and contrast between local data augmentations and local feature augmentations used in \method ($\rm{L_d2L_f}$).

The results are shown in Table \ref{table:ablation_mode}.
With multi-crop, DINO \cite{dino_2021} (81.1\%) improves accuracy by 3.0\% compared to DINO without multi-crop baseline.
\method (84.8\%) further improves accuracy by 3.7\% by introducing a contrast between local data augmentation and local feature augmentation.
Some attention maps of \method and DINO are shown in Fig. \ref{fig:exp_attention}.
\method can more accurately focus on semantics-consistent regions between global view and local views, while DINO is easily influenced by background.

We also add multi-layer feature contrastive learning on DINO.
The result in Table \ref{table:ablation_mode} (the ``DINO w ML" row) verifies the improvements of \method are not totally owed to multi-layer contrast.

In addition, we directly crop the corresponding region of local augmentation on the feature map of global augmentation for contrastive learning.
As shown in Table \ref{table:ablation_mode}, this variant (ROIAlign) of \method also outperforms the DINO baseline, which shows that the directly cropped features are also beneficial for contrastive learning.
And the ROIAlign variant of \method achieves lower accuracy than \method, demonstrating that the soft feature search in \method is better than the hard ROIAlign since ROIAlign may damage the continuous semantic context of the feature map.

Further, we also test the performance of \method under the setting without multi-crop.
That is, the feature search is conducted between two global data augmentations.
We term this contrast mode as $\rm{G_d2G_f}$.
As shown in the ``\method w/o MC" row, \method also improves the performance compared to its baseline (the ``DINO w/o MC" row), which proves that \method is also helpful in solving the semantic inconsistency caused by other augmentations, not only the multi-crop augmentation strategy.

\setlength{\tabcolsep}{0.4mm}
\begin{table}[t]
\begin{center}
\begin{tabular}{l|cccc|cc}
\toprule
Contrast Mode & $\rm{G_d2G_d}$ & $\rm{L_d2G_d}$ & $\rm{L_d2L_f}$ & $\rm{G_d2G_f}$ & $k$-NN & LP \\
\hline
DINO w/o MC & $\checkmark$ & & & & 78.1 & 83.7 \\
DINO & $\checkmark$ & $\checkmark$ & & & 81.1 & 87.0 \\
DINO w ML & $\checkmark$ & $\checkmark$ & & & 82.2 & 87.4 \\
\method & $\checkmark$ & $\checkmark$ & $\checkmark$ & & 84.8 & 89.2 \\
ROI Crop & $\checkmark$ & $\checkmark$ & $\checkmark$ & & 83.9 & 88.1 \\
\method w/o MC & & & & $\checkmark$ & 79.7 & 86.3 \\
\bottomrule
\end{tabular}
\end{center}
\setlength{\abovecaptionskip}{0.05cm}
\caption{Influence of different contrast modes. MC, ML, and LP denote multi-crop, multi-layer, and linear probing, respectively.}
\label{table:ablation_mode}
\end{table}

\begin{figure}[t]
\centering
\includegraphics[width=0.95\columnwidth]{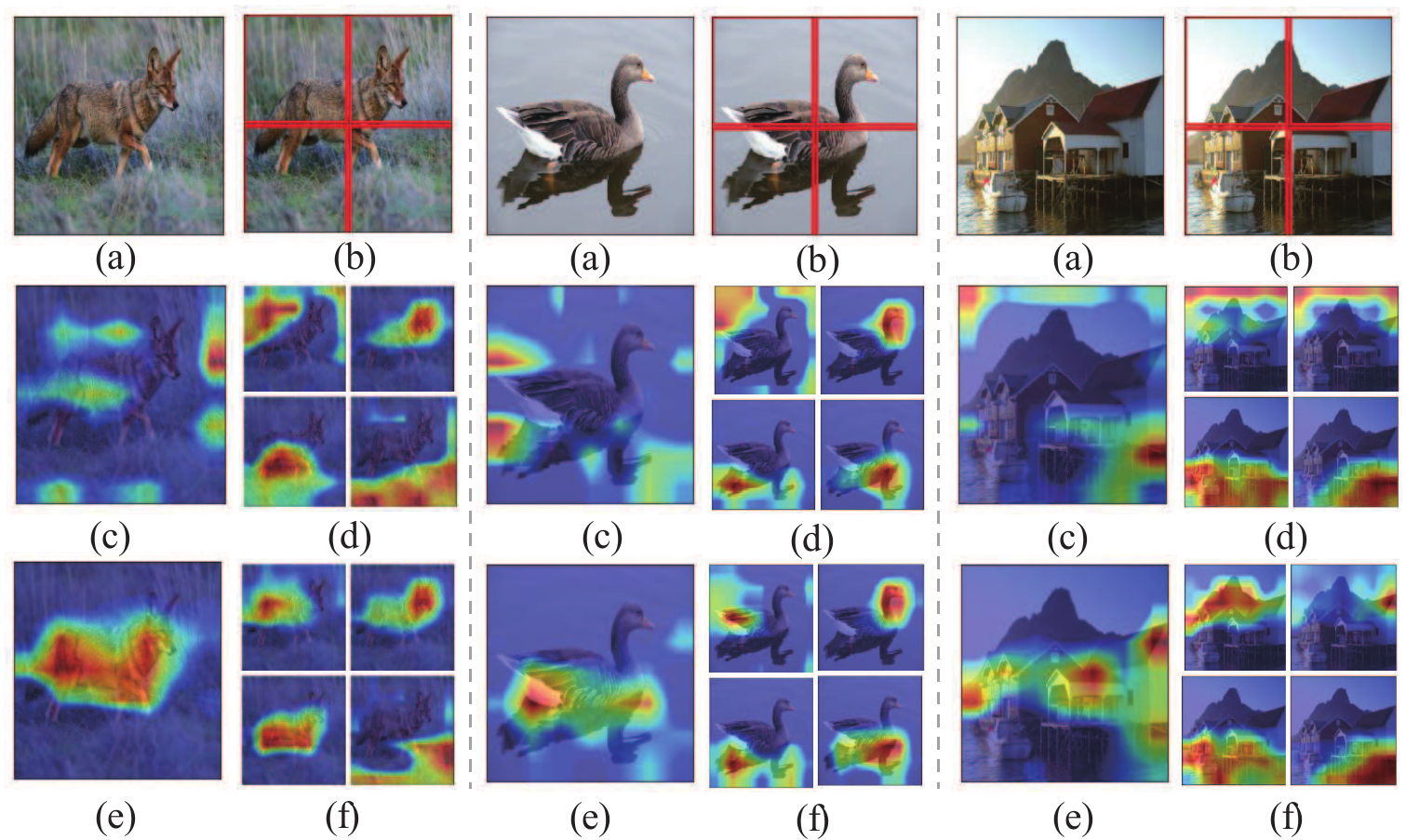}
\setlength{\abovecaptionskip}{0.05cm}
\caption{Attention maps of \method (the third row) compared with DINO \cite{dino_2021} (the second row).
In each example, (a) shows a global image, and its four local images in (b) are constructed by $2\times2$ jigsaw.
(d) and (f) show the attention maps that highlight the semantics-consistent regions between the local images in (b) and the global image in (a).
They are obtained by multiplying the globally average pooled feature maps from the encoder (Res4) of the local images in (b) with the feature map (Res4) of the global image in (a).
And the encoder is the trained DINO ResNet50 model and \method ResNet50 model in (d) and (f), respectively.
(c) and (e) show the mean attention maps of DINO and \method respectively, which are obtained by multiplying the mean globally average pooled feature map of the four local images in (b) with feature map of the global image in (a).
}
\label{fig:exp_attention}
\end{figure}

\noindent
\textbf{Influence of Multi-Layer Contrast.}
The influence of the feature layer that is used for feature search is analyzed.
The $Res2$, $Res3$ and $Res4$ in the ResNet50 \cite{resnet_2016} backbone are evaluated.
As shown in Table \ref{table:ablation_layer}, the performance improves with the increase of feature layer numbers, which demonstrates that conducting feature search on more layers is helpful for representation learning.

\setlength{\tabcolsep}{6pt}
\begin{table}[!t]
\begin{center}
\begin{tabular}{ccc|cc}
\toprule
Res2 & Res3 & Res4 & $k$-NN & Linear Probing \\
\hline
$\checkmark$ & & & 82.0 & 86.1 \\
$\checkmark$ & $\checkmark$ & & 84.3 & 87.5 \\
$\checkmark$ & $\checkmark$ & $\checkmark$ & 84.8 & 89.2 \\
\bottomrule
\end{tabular}
\end{center}
\setlength{\abovecaptionskip}{0.05cm}
\caption{Influence of different feature augmentation layer.}
\label{table:ablation_layer}
\end{table}

\begin{figure}
\centering
\includegraphics[width=0.8\columnwidth]{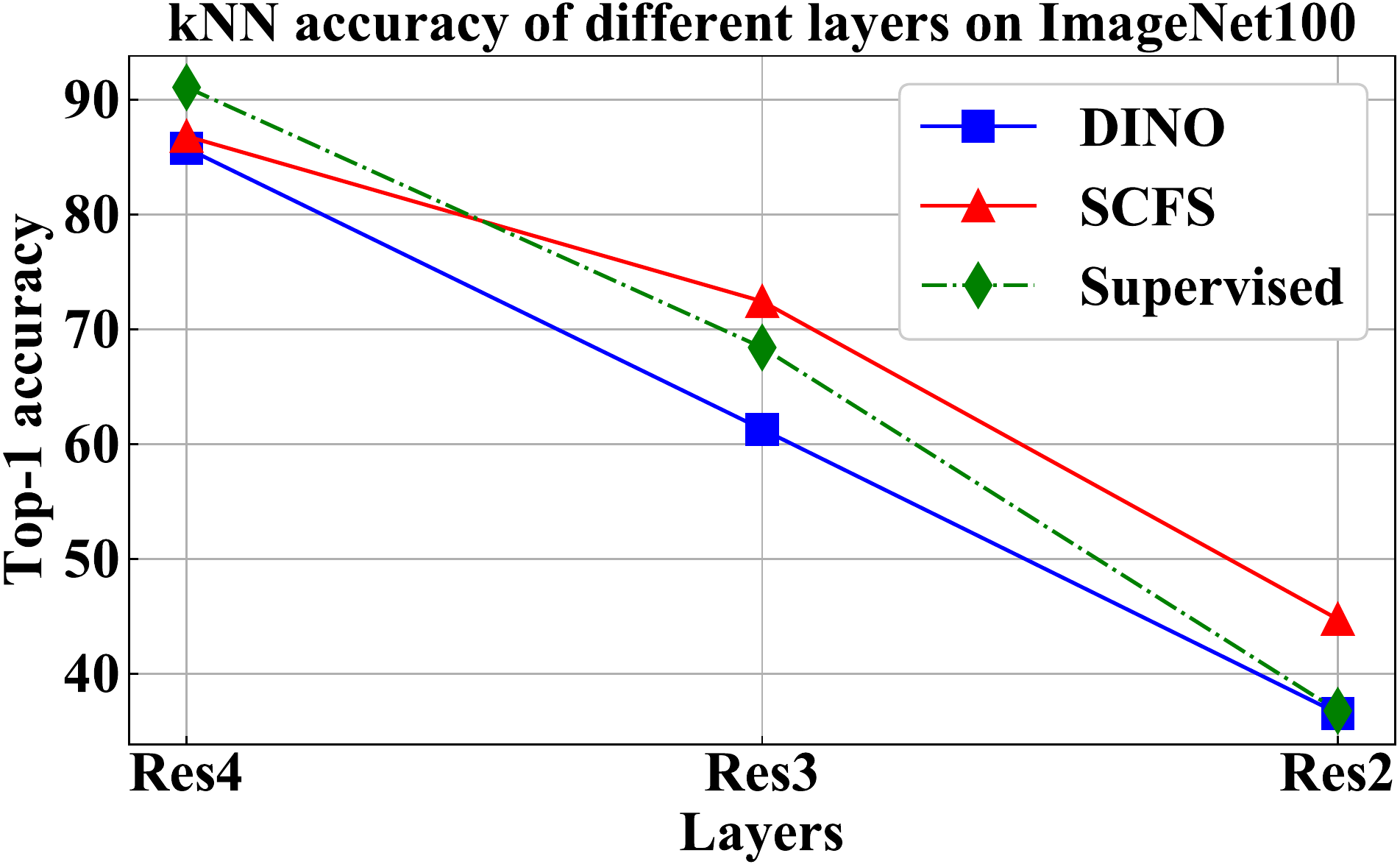}
\setlength{\abovecaptionskip}{0.05cm}
\caption{The $k$-NN accuracy of features from different layers.}
\label{fig:exp_layersknn}
\end{figure}

Further, we evaluate the $k$-NN accuracy using feature maps from different layers to observe the influence of feature search on the representation of middle layers.
We also choose the features extracted by the $Res2$, $Res3$ and $Res4$ layer of ResNet50.
The results are shown in Fig. \ref{fig:exp_layersknn}.
Compared with DINO \cite{dino_2021}, \method achieves better performance with features from all middle layers on ImageNet100, which verifies that enhancing the semantic consistency can improve the semantic representation of shallow layers.
Compared with supervised learning, \method model has higher performance on res2 and res3 layer, which shows that \method is more advantageous in the shallow layer feature representation.



\begin{figure}[!h]
\centering
\includegraphics[width=0.8\columnwidth]{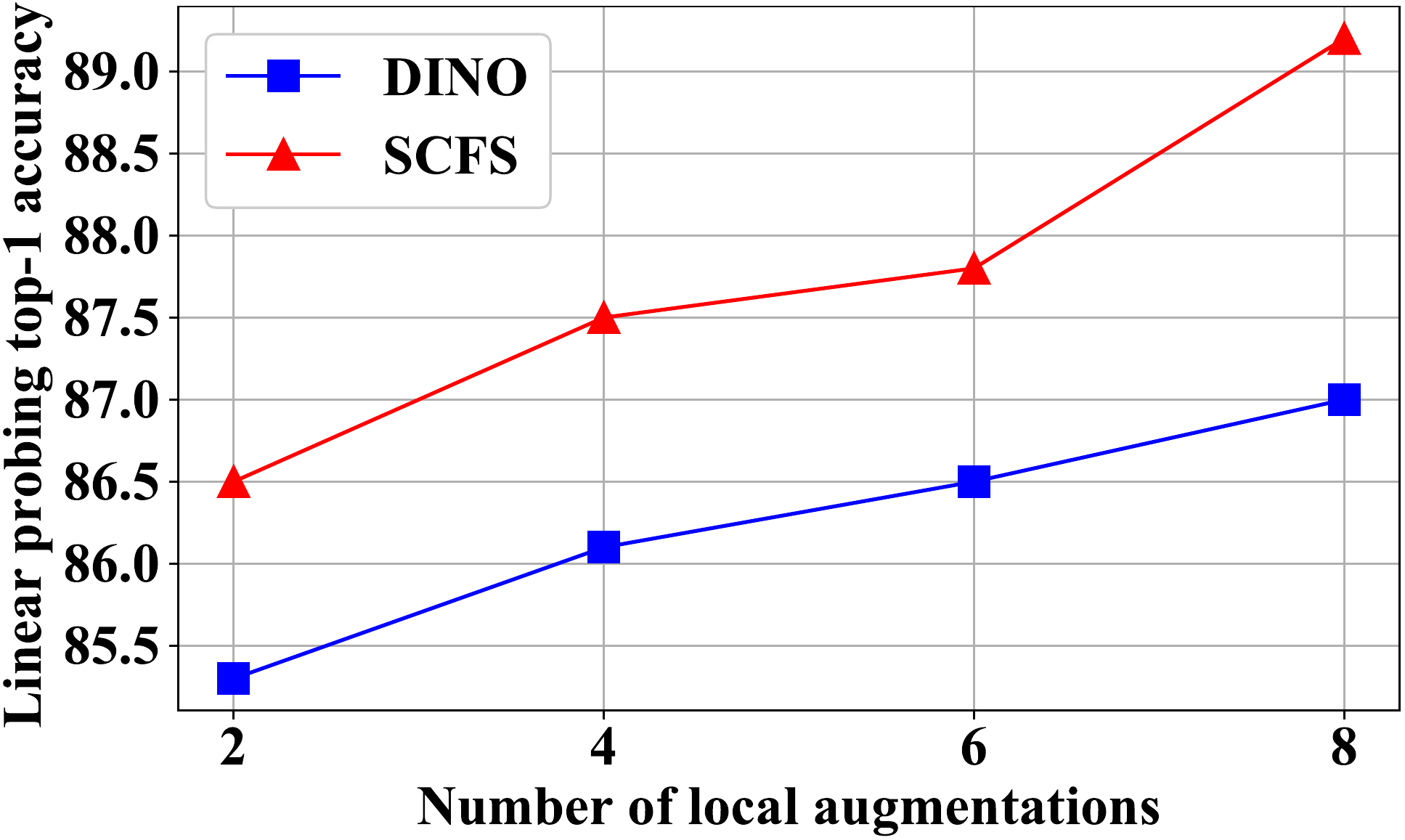}
\setlength{\abovecaptionskip}{0.05cm}
\caption{Influence of local augmentation number.}
\label{fig:localnumber}
\end{figure}

\noindent
\textbf{Influence of Local Augmentation Number.}
In this part, we analyze the performance difference with the change of local augmentation numbers.
The results are shown in Fig. \ref{fig:localnumber}.
The performance of DINO and \method is steadily improved when adding more local augmentations for contrast.
In addition, \method improves the performance under different local augmentation numbers, which demonstrates that semantics-consistent feature search is helpful to alleviate the influence of semantics inconsistent data augmentations.

\noindent
\textbf{Experiments on Other Backbones.}
In this part, we conduct experiments on other backbones to further evaluate the effectiveness of \method.
Apart for the default Reset50 used in other experiments, ResNet101 and Vision Transformer (ViT-S and ViT-B) are tested.
The results are shown in Table \ref{table:backbones}.
\method achieves significant improvement on different backbones compared to its baseline DINO, which demonstrates that \method is applicable to different backbones.

\setlength{\tabcolsep}{3pt}
\begin{table}[]
\centering
\begin{tabular}{cccc|cc}
\hline
Method & Backbone & Batch Size & Epochs & $k$-NN & LP \\ \hline
\toprule
DINO & R101 & 256 & 200 & 81.0 & 86.3 \\
\rowcolor{Gray}\textbf{\method} & R101 & 256 & 200 & \textbf{85.1} & \textbf{88.3} \\ \hline
DINO & ViT-S & 256 & 200 & 75.0 & 80.4 \\
\rowcolor{Gray}\textbf{\method} & ViT-S & 256 & 200 & \textbf{76.3} & \textbf{81.0} \\ \hline
DINO & ViT-B & 256 & 200 & 76.2 & 80.7 \\
\rowcolor{Gray}\textbf{\method} & ViT-B & 256 & 200 & \textbf{77.2} & \textbf{82.3} \\
\bottomrule
\end{tabular}
\setlength{\abovecaptionskip}{0.1cm}
\caption{Experiments on other backbones. LP denotes linear probing.}
\label{table:backbones}
\end{table}

\section{Conclusions}
\label{sec:conclusion}

In this study, we aim to alleviate the problem of unmatched semantic alignment in current contrastive learning by expanding the augmentations from data space to feature space.
The proposed semantics-consistent feature search (\method) adaptively searches semantics-consistent local features between different views for contrast, while suppressing irrelevant local features during pre-training.
It conducts contrast learning between feature augmentation and data augmentation.
The experimental results demonstrate that \method can learn to focus on meaningful object regions and effectively improve the performance of self-supervised learning.
The feature search procedure in \method is learnable parameter-free.
We will utilize the self-attention mechanism in Transformer to perform the feature search procedure to further boost its performance in future work.

{\small
\bibliographystyle{ieee_fullname}
\bibliography{SCFS}
}

\clearpage
\renewcommand\thefigure{S\arabic{figure}}
\renewcommand\thetable{S\arabic{table}}
\renewcommand\theequation{S\arabic{equation}}
\setcounter{equation}{0}
\setcounter{table}{0}
\setcounter{figure}{0}
\setcounter{section}{0}
\renewcommand\thesection{\Alph{section}}

\section*{Appendix}

\section{Hyper-parameters Setting}
\label{sec:apd_hyperparameters}
During the pretraining procedure, we follow the most hyper-parameters setting of DINO \cite{dino_2021}.
The SGD optimizer is used and the learning rate is linearly warmed up to its base value during the first 10 epochs.
The base learning rate is set according to the linear scaling rule: $lr=0.1\times batchsize / 256$.
After the warm-up procedure, the learning rate is decayed with a cosine schedule \cite{cosine_decay_2016}.
The weight decay is set to $1e-4$.
For the temperatures, $\tau$ is set to 0.1, and a linear warm-up from 0.04 to 0.07 is set to $\tau'$ during the first 50 epochs.
Following DINO \cite{dino_2021}, the centering operation is applied to the output of the momentum encoder to avoid collapse.
For data augmentation, the global augmentations consist of random cropping (with a scale of 0.14-1), resizing to $224 \times 224$, random horizontal flip, gaussian blur, and color jittering.
And the local augmentations consist of random cropping (with a scale of 0.05-0.14), resizing to $96 \times 96$, random horizontal flip, gaussian blur, and color jittering.
2 global views with $N=8$ local views are the default setting of augmentation.

During the linear probing procedure, we evaluate the representation quality with a linear classifier.
The linear classifier is trained with the SGD optimizer and a batch size of 1024 for 100 epochs on ImageNet.
Weight decay is not used.
For data augmentation, only random resizes crops and horizontal flips are applied.

\section{Projection Head}
\label{sec:apd_projectionhead}
\begin{figure}[h]
\centering
\includegraphics[width=0.95\columnwidth]{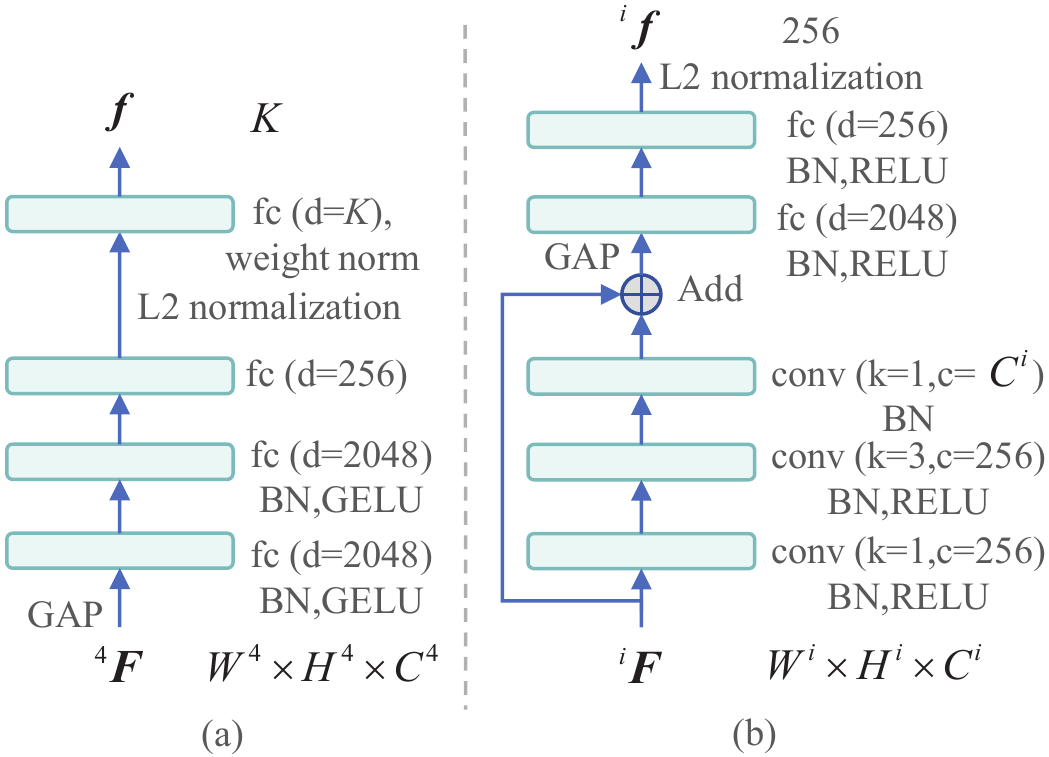}
\caption{Architecture of the projection heads in \method. (a) projection head for the contrast between data augmentations; (b) projection head for feature search.
}
\label{fig:appendix_heads}
\end{figure}

There are two kinds of projection heads in \method.
The projection head for the contrast between data augmentations consists of a four-layer MLP with the same architecture as DINO \cite{dino_2021}.
As shown in \cref{fig:appendix_heads} (a), the hidden layers are with 2048 dimension and are with gaussian error linear units (GELU) activations.
After the MLP, a $L_2$ normalization and a weight normalized FC layer with $K$ ($K=65536$) dimension are applied.

The projection head for feature search consists of three convolutional layers and two FC layers.
The detailed architecture is shown in \cref{fig:appendix_heads} (b).
To make the feature search loss easy to backward, the residual connection is applied to the three convolutional layers.
After global-averaged pooling, two FC layers are applied to project features to the output dimension.
Note that the output dimension is set to $256$, which achieves good performance in all the experiments.

\section{Training Time}
We test the training times on a machine with 8 NVIDIA GeForce RTX 2080Ti GPUs.
As shown in \cref{tab_time}, compared to the baseline DINO \cite{dino_2021}, the extra computational time of \method increases by 30\%.

\begin{table}[]
\centering
\caption{Training Time.}
\label{tab_time}
\begin{tabular}{c|c|c|c}
\hline
Method & Batch Size & Epochs & Time \\ \hline
\toprule
DINO & 256 & 200 & 147h \\
\method & 256 & 200 & 192h \\
\bottomrule
\end{tabular}
\end{table}

\section{More Visualization Results}
\begin{figure*}[h]
\centering
\includegraphics[width=0.8\linewidth]{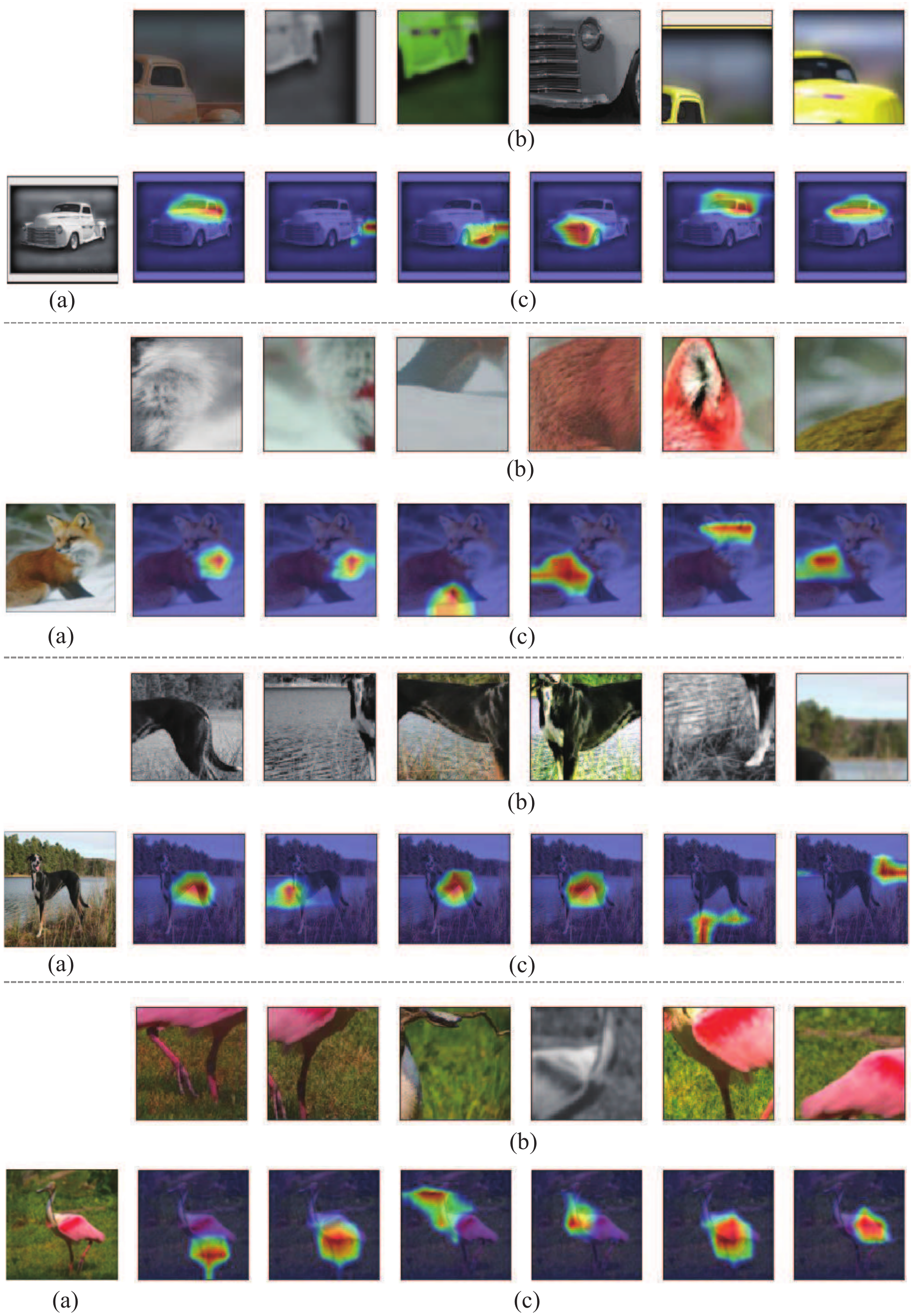}
\caption{
Attention maps of \method between local images and corresponding global image.
In each example, (a) shows a global image, (b) shows six local augmentations of the global image, and (c) shows the attention maps that highlight the semantics-consistent regions between the local images in (b) and the global image in (a), which are obtained by multiplying the globally average pooled feature maps from the encoder (Res4) of the local images in (b) with the feature map (Res4) of the global image in (a).
}
\label{fig:appendix_attention}
\end{figure*}

We visualize the attention maps of \method between local images and corresponding global image.
As shown in \cref{fig:appendix_attention}, \method can accurately focus on semantics-consistent regions between global images and local images. According to the different semantic concepts inputs, consistent semantic information can be searched on the global feature.

\begin{figure*}[!h]
\centering
\includegraphics[width=0.8\linewidth]{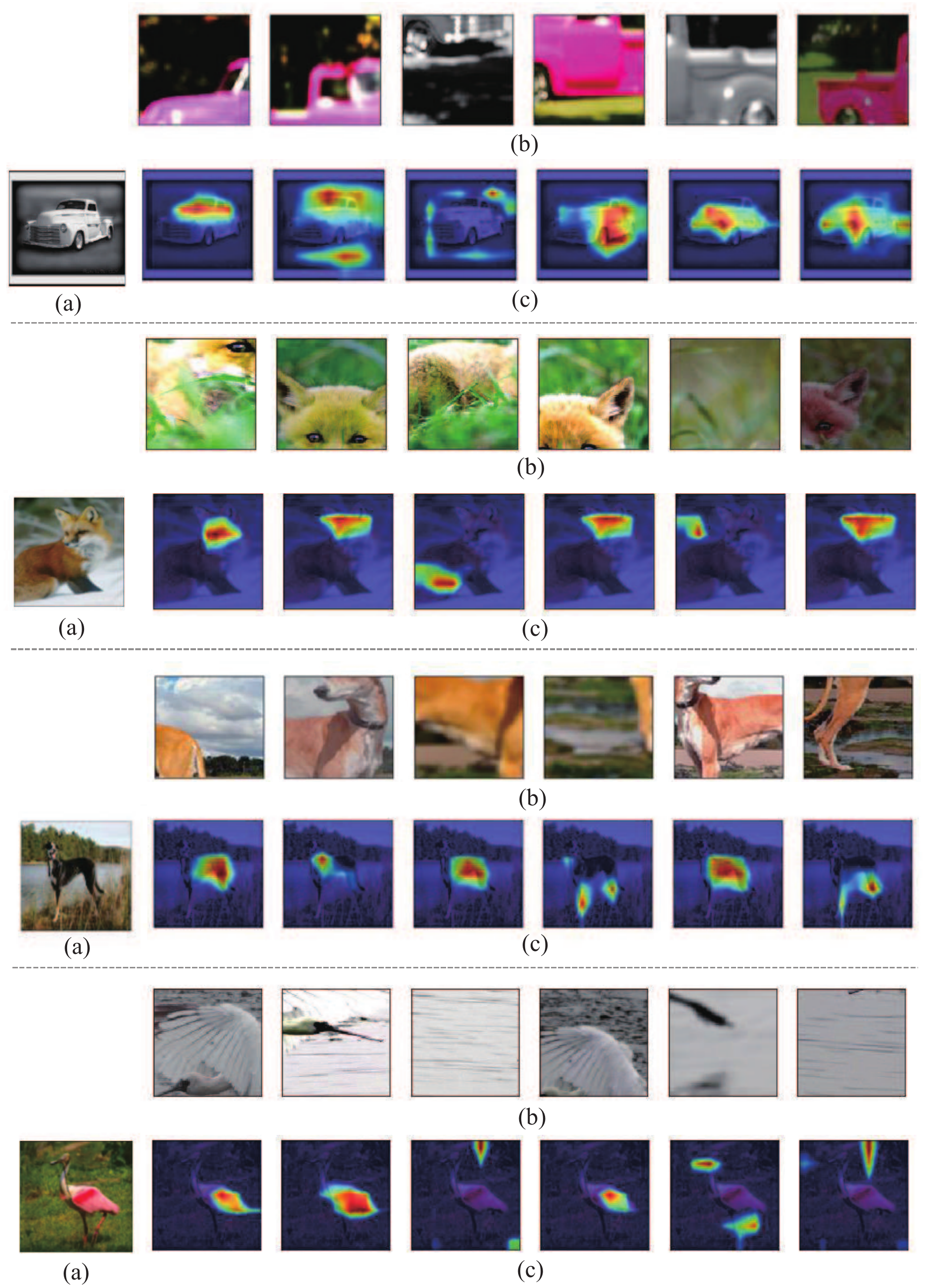}
\caption{
Attention maps of \method between local images and another image that contains objects with the same category.
In each example, (a) shows an image that contains objects with the same category in (b), (b) shows six local augmentations of a global image, and (c) shows the attention maps that highlight the semantics-consistent regions between the local images in (b) and the image in (a), which are obtained by multiplying the globally average pooled feature maps from the encoder (Res4) of the local images in (b) with the feature map (Res4) of the image in (a).
}
\label{fig:appendix_attention_multi}
\end{figure*}

Furthermore, we also visualize the attention maps between local images and another image that contains objects with the same category.
As shown in \cref{fig:appendix_attention_multi}, the attention maps show that the semantics-consistent regions between different images are also activated. When the background images are input, the global images are no longer activated incorrectly, which achieves the contrastive noise mitigation and demonstrates the effectiveness of \method.

\end{document}